\documentclass[sigconf]{acmart}

\usepackage{algorithmic}
\usepackage{graphicx}
\usepackage{textcomp}
\usepackage{xcolor}
\usepackage[disable]{todonotes}
\usepackage{multirow}
\usepackage{listings}
\usepackage{subcaption}
\usepackage{xcolor}
\usepackage{enumitem}

\setlist{nolistsep}
\definecolor{codegreen}{rgb}{0,0.6,0}
\definecolor{codegray}{rgb}{0.5,0.5,0.5}
\definecolor{codepurple}{rgb}{0.58,0,0.82}
\definecolor{backcolour}{rgb}{0.95,0.95,0.92}

\lstdefinestyle{mystyle}{
  backgroundcolor=\color{backcolour}, commentstyle=\color{codegreen},
  keywordstyle=\color{magenta},
  numberstyle=\tiny\color{codegray},
  stringstyle=\color{codepurple},
  basicstyle=\ttfamily\footnotesize,
  breakatwhitespace=false,         
  breaklines=true,                 
  captionpos=b,                    
  keepspaces=true,                 
  numbers=left,                    
  numbersep=5pt,                  
  showspaces=false,                
  showstringspaces=false,
  showtabs=false,                  
  tabsize=2
}

\colorlet{punct}{red!60!black}
\definecolor{background}{HTML}{EEEEEE}
\definecolor{delim}{RGB}{20,105,176}
\colorlet{numb}{magenta!60!black}

\lstdefinelanguage{json}{
    basicstyle=\normalfont\ttfamily,
    numbers=left,
    numberstyle=\scriptsize,
    stepnumber=1,
    numbersep=8pt,
    showstringspaces=false,
    breaklines=true,
    frame=lines,
    backgroundcolor=\color{background},
    literate=
     *{0}{{{\color{numb}0}}}{1}
      {1}{{{\color{numb}1}}}{1}
      {2}{{{\color{numb}2}}}{1}
      {3}{{{\color{numb}3}}}{1}
      {4}{{{\color{numb}4}}}{1}
      {5}{{{\color{numb}5}}}{1}
      {6}{{{\color{numb}6}}}{1}
      {7}{{{\color{numb}7}}}{1}
      {8}{{{\color{numb}8}}}{1}
      {9}{{{\color{numb}9}}}{1}
      {:}{{{\color{punct}{:}}}}{1}
      {,}{{{\color{punct}{,}}}}{1}
      {\{}{{{\color{delim}{\{}}}}{1}
      {\}}{{{\color{delim}{\}}}}}{1}
      {[}{{{\color{delim}{[}}}}{1}
      {]}{{{\color{delim}{]}}}}{1},
}
\AtBeginDocument{%
  \providecommand\BibTeX{{%
    \normalfont B\kern-0.5em{\scshape i\kern-0.25em b}\kern-0.8em\TeX}}}

\setcopyright{acmcopyright}
\copyrightyear{2023}
\acmYear{2023}
\acmDOI{XXXXXXX.XXXXXXX}

\acmConference[Correctness 2023]{Seventh International Workshop on Software Correctness for HPC Applications}{November 12,
  2023}{Denver, CO}
\setcopyright{rightsretained} 
\begin{document}
\copyrightyear{2023} 
\acmYear{2023} 

\acmConference[SC-W 2023]{Workshops of The International Conference on High Performance Computing, Network, Storage, and Analysis}{November 12--17, 2023}{Denver, CO, USA}
\acmBooktitle{Workshops of The International Conference on High Performance Computing, Network, Storage, and Analysis (SC-W 2023), November 12--17, 2023, Denver, CO, USA}
\acmDOI{10.1145/3624062.3624088}
\acmISBN{979-8-4007-0785-8/23/11}
\title{Data Race Detection Using Large Language Models}

\author{Le Chen}
\email{lechen@iastate.edu}
\affiliation{%
  \institution{Iowa State University}
  \city{Ames}
  \state{IA}
  \country{USA}
}
\affiliation{%
  \institution{Lawrence Livermore National Laboratory}
  \city{Livermore}
  \state{CA}
  \country{USA}}


\author{Xianzhong Ding} 
\email{xding5@ucmerced.edu}
\affiliation{%
  \institution{University of California, Merced}
  \city{Merced}
  \state{CA}
  \country{USA}
  }
\affiliation{%
  \institution{Argonne National Laboratory }
  \city{Lemont}
  \state{IL}
  \country{USA}}

\author{Murali Emani} 
\email{memani@anl.gov}
\affiliation{%
  \institution{Argonne National Laboratory }
  \city{Lemont}
  \state{IL}
  \country{USA}}

\author{Tristan Vanderbruggen, \\ Pei-Hung Lin, Chunhua Liao} 
\email{{vanderbrugge1
,lin32,liao6}@llnl.gov}
\affiliation{%
  \institution{Lawrence Livermore National Laboratory}
  \city{Livermore}
  \state{CA}
  \country{USA}}

%

\renewcommand{\shortauthors}{Chen, et al.}

\begin{abstract}
Large language models (LLMs) are demonstrating significant promise as an alternate strategy to facilitate analyses and optimizations of high-performance computing programs, circumventing the need for resource-intensive manual tool creation. In this paper, we explore a novel LLM-based data race detection approach combining prompting engineering and fine-tuning techniques. We create a dedicated dataset named DRB-ML, which is derived from DataRaceBench, with fine-grain labels showing the presence of data race pairs and their associated variables, line numbers, and read/write information. DRB-ML is then used to evaluate representative LLMs and fine-tune open-source ones. 
Our experiment shows that LLMs can be a viable approach to data race detection. However, they still cannot compete with traditional data race detection tools when we need detailed information about variable pairs causing data races. 

\end{abstract}






\maketitle

\section{Introduction} 


The advent of many-core and GPU-accelerated systems has fostered the need for threaded programming models, such as OpenMP and CUDA, to exploit intra-node parallelism effectively. However, a perennial challenge accompanying these programming models is the risk of data races. Data races transpire when two or more threads access the same memory location simultaneously in a conflicting manner, without sufficient synchronization, with at least one of these accesses involving a write operation. This type of bug induces unpredictable behaviors in code, meaning they may not consistently manifest each time the code is executed, thereby exacerbating their detection and resolution difficulties.

Various tools, such as Intel Inspector~\cite{Intel_Inspector} and ThreadSanitizer~\cite{THREADSANITIZER}, have been developed to assist developers in responding to the challenges associated with data race detection in multithreaded programs.  
Due to the fast evolution of parallel programming, these tools need to be constantly re-evaluated and manually updated to support new language features and code patterns.
A dataset explicitly designed for data race analysis, named DataRaceBench (DRB)~\cite{liao2017dataracebench}, was introduced for evaluating the performance and effectiveness of various detection tools and methodologies.

In recent years, the realm of machine learning has been illuminated by significant breakthroughs, particularly the emergence of Large Language Models (LLMs). Harnessing the power of deep learning \cite{ding2019octopus}, LLMs have demonstrated their proficiency in comprehending and generating human-like text from provided prompts. Given the remarkable ability of LLMs to understand and generate text, they hold immense potential in the field of Programming Language Processing (PLP) tasks~\cite{chen2021evaluating, zan2023large, chen2023learning, lei2023creating}. This includes but is not limited to tasks such as code analysis, generation, and bug detection, showcasing an exciting expansion beyond the traditional confines of Natural Language Processing (NLP).


Building on their numerous applications in PLP, LLMs present potential for deployment in the specialized area of data race detection. We envisage that a fine-tuned LLM for data race detection could discern common patterns and contexts that precipitate data races. When applied to unfamiliar code, it could forecast potential data race conditions, thus equipping developers with a proactive alert mechanism to mitigate possible risks. Compared to traditional static or dynamic analysis methods, if adequately trained, machine learning approaches impose minimal human labor and runtime overhead and can effectively respond to a broad spectrum of data race patterns. Furthermore, with their innate capability to generate human-like text, LLMs could facilitate detailed explanations about detected data race conditions. Such insights could guide developers in discerning the underlying causes of these bugs and subsequently aid in devising more effective resolution strategies.

The application of Large Language Models in data race detection represents a pioneering field of research. Our work commences by generating a comprehensive dataset distinctly labeled with explicit data dependencies and data race information, emphasizing our commitment to ensuring the accuracy and dependability of the model's output. Subsequently, we propose novel experimental approaches for data race detection using LLMs.
The following contributions distinguish this paper:
\begin{enumerate}[leftmargin=*]
\item We have derived an innovative dataset from DataRaceBench for machine learning training and large language model fine-tuning.
\item We extensively evaluate several prominent LLMs using various prompting techniques.
\item We fine-tune LLMs for the explicit task of data race detection, thereby enhancing their predictive accuracy. 
\item A detailed comparative study between traditional data race detection tools and LLM-based methods, highlighting their strengths and weaknesses.
\end{enumerate}



\section{Background} 
In this section, we provide an overview of large language models and their application in the context of programming language processing. We also introduce DataRaceBench. 

\subsection{Large Language Models}

Large Language Models (LLMs) are large-sized machine learning models specifically designed to perform various natural language processing tasks, such as analyzing and generating text, answering questions in a conversational manner, and translating text from one language to another. Previous work ~\cite{llmsurvey} observed that large-sized pre-trained language models exhibit behaviors distinct from smaller ones (e.g., 330M-parameter BERT and 1.5B-parameter GPT-2) and show surprising abilities (called emergent abilities). Large language models have emerged as revolutionary tools in machine learning. The surging popularity of LLMs can be attributed to their versatile applicability and unparalleled performance in diverse tasks. LLMs have consistently outperformed traditional models, from enhancing natural language processing applications like sentiment analysis ~\cite{zhang2023sentiment} and chatbots \cite{kasneci2023chatgpt} to aiding researchers in content generation, summarization, and translation ~\cite{karpinska2023large, lei2023boosting}. 

Despite their inherent proficiency in context-dependent learning, pre-trained LLMs often require additional training or fine-tuning to perform specialized or novel tasks. This process allows the models to adapt to specific problem domains, thereby improving their performance and relevance in a given context.

%


Integrating Natural Language Processing (NLP) techniques in Programming Language Processing (PLP) tasks has sparked substantial interest. With applications that extend to code summarization, code generation, and code similarity analysis~\cite{chen2022multi,flynn2022finding}, this emerging field has witnessed the successful deployment of traditional language models, underscoring the viability and potential of this approach~\cite{devlin2018bert}.

A remarkable advancement in this domain is the adaptation of transformer-based language models for PLP tasks. Representative models, such as CodeBERT~ \cite{feng2020codebert} and CodeT5~\cite{wang2021codet5}, epitomize this trend. These models leverage a transformer architecture and are trained on a wide array of programming languages to facilitate an extensive spectrum of programming-related tasks.

In the context of Large Language Models for Code (Code LLMs), several works~\cite{shen2023pangu, chen2023lm4hpc, ding2023hpcgpt} have explored pre-trained LLMs, either general-purpose or task-specific, for PLP tasks. In this study, we focus on four representative LLMs: GPT-3.5-turbo, GPT-4, Llama2-7b, and StarChat-beta (StarChat-$\beta$), which demonstrate diverse capabilities and applications in the sphere of code analysis and generation.

GPT-3.5-turbo ~\cite{brown2020language}, engineered by OpenAI, is a state-of-the-art language model capable of generating human-like text and comprehending nuanced prompts. For our research, we employ the 16k version of the model, accommodating up to 16384 input tokens.
Succeeding GPT-3.5-turbo, GPT-4~\cite{OpenAI2023GPT4TR} marks OpenAI's latest and most powerful offering. While GPT-4 retains the fundamental architecture of its predecessor, it capitalizes on expanded training across a diverse range of internet text, thereby enhancing both the model's size and capabilities.
Llama2~\cite{touvron2023llama} is one of the latest models released by Meta. As a substantial and robust language model, it demonstrates particular strength in tasks requiring deep understanding and information synthesis. 
Based on StarCoder~\cite{li2023starcoder}, StarChat constitutes a series of GPT-style models explicitly crafted for code-related tasks. Their base models are 15.5B parameter models trained on 80+ programming languages. StarChat-beta is the second model in this series with 16B parameters. 

An LLM-based approach offers distinct advantages, including the capacity to automatically capture common patterns across similar languages and to avoid the need for manual tool development for individual languages. Compared with previous machine learning-based approaches ~\cite{tehranijamsaz2021deeprace}, LLMs can be continually fine-tuned on new data, adapting to new domains or specific tasks while still retaining their broad capabilities.  However, the potential of LLMs in the realm of data race detection has yet to be fully explored. This research aims to probe into and unravel their capabilities in this domain.

\subsection{Data Race Detection and DataRaceBench}
Prominent techniques for data race detection leverage two major techniques: static and dynamic analysis.  Static analysis tools such as Locksmith~\cite{pratikakis2006locksmith}, RELAY~\cite{voung2007relay}, and ompVerify~\cite{basupalli2011ompverify} inspect program source code or intermediate representations (IRs) to reveal potential data races through control flow and data dependency analysis. 
On the other hand, dynamic analysis tools such as Inter Inspector~\cite{Intel_Inspector} and ThreadSanitizer ~\cite{THREADSANITIZER} inspect program behavior during execution by instrumenting code to monitor memory accesses in real time.  This is achieved by instrumenting the code to observe memory access in real time. Techniques under dynamic analysis, such as lockset-based and happens-before-based detection, often yield better accuracy in data race detection.

Despite the accurate output, dynamic analysis methods introduce runtime overhead and will likely miss certain race conditions that are hard to trigger during testing. 
Static analysis methods, in contrast, analyze the source code without executing it and generally offer faster results. Static analysis is advantageous in identifying race conditions that might not manifest during dynamic testing. 
With the massive number of threads available in the latest computing architectures, the interest in static analysis is growing to complement dynamic analysis in performing race detection for modern systems.

Hybrid approaches that exploit both static and dynamic analyses become promising for discerning potential data races with increased fidelity.
Nowadays, with the breakthrough in machine learning technology  \cite{ding2020mb2c, ding2022drlic}, methodologies with machine learning have gained traction, leveraging pattern recognition in program behavior or applying LLMs to enable data race detection.

DataRaceBench (DRB) is an open-source benchmark suite methodically and quantitatively designed to evaluate data race detection tools. It is particularly oriented towards the context of OpenMP, a widely used parallel programming model for multithreaded applications. More specifically, DRB contains microbenchmark programs both with and without data races, which are either manually crafted, derived from actual scientific applications, or generated as automatic optimization variants.

Despite its effective labeling system for collected microbenchmarks, DRB lacks a structured dataset specifically designed for machine learning training and evaluation. There is a clear demand for a well-curated dataset comprising prompt-response pairs, which is crucial for the fine-tuning process of LLMs. Such a dataset, tailor-made for machine learning applications, could significantly enhance the performance and efficacy of data race detection methodologies.

\section{Approach} 
In this section, we elaborate on our approach by combining two principal routes designed to exploit the capabilities of Large Language Models for novel tasks. First, we evaluate three strategies for data race detection with LLMs. Second, we fine-tune two open-source LLMs for data race detection and identification of data race variable pairs. As shown in Figure 1, our approach relies on the proposed dataset, DataRaceBench-ML. This section uses several popular Large Language Models, such as GPT, Llama2, and StarChat-beta. 

\begin{figure}[h]
  \centering
  \includegraphics[width=\linewidth]{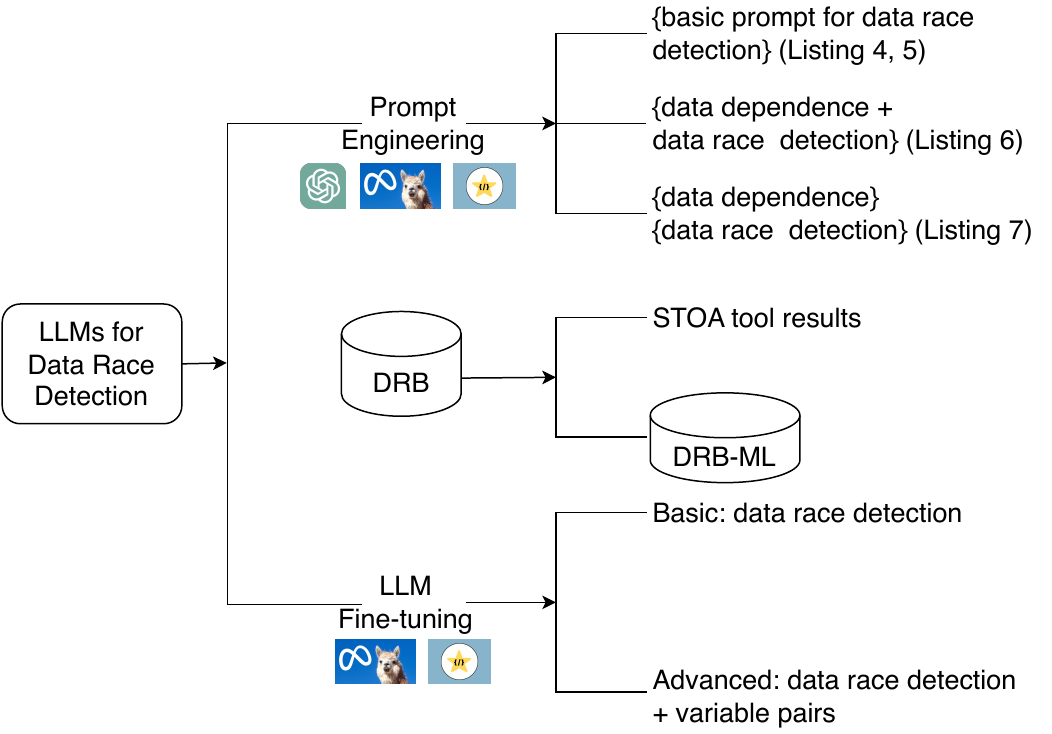}
  \caption{Overview of Our Approach for Data Race Detection using Large Language Models: GPT-3.5-turbo, GPT-4, Llama2, and StarChat-beta.}
  \label{fig:pipe}
\end{figure}



\subsection{DataRaceBench-ML Dataset}
\label{sec:drbml}
The quality of datasets is essential for the success of any machine-learning approach. They determine the accuracy and robustness of a model's predictions. We processed the existing DataRaceBench V1.4.1 (DRB) to generate a new dataset, DataRaceBench-ML (DRB-ML), to facilitate efficient ML model training, fine-tuning, and evaluation. Each C/C++ microbenchmark from the DRB results in an entry in DRB-ML. Consequently, DRB-ML consists of 201 JSON files storing various key-value pairs - a direct correlation to the number of code snippets in the original DRB dataset. 

\begin{table*}[h]
\caption{Keys and Values in DRB-ML }
\label{tab:drbml_keys}
\begin{tabular}{|l|c|l|}
\hline
Keys              & Value type & Description                                                                                \\ \hline
ID                & int        & A unique index number starting from 1.                                                     \\ \hline
name              & str        & The original filename of the DRB file.                                                     \\ \hline
DRB\_code         & str        & The original code present in DRB microbenchmarks.                                          \\ \hline
trimmed\_code     & str        & The DRB\_code with all comments removed.                                                   \\ \hline
code\_len         & int        & An integer value representing the string length of the trimmed code.                       \\ \hline
data\_race        & int        & The presence of a data race is indicated by 1, and its absence by 0.                       \\ \hline
data\_race\_label & str        & This label indicates the type of data race condition (race-yes or race-no) that DRB marks. \\ \hline
var\_pairs &
  [str] &
  \begin{tabular}[c]{@{}l@{}}This is a list of pairs of variables associated with a data race. It is empty when "data\_race" is 0. \\ Each item follows the format [VAR0, VAR1], where VAR1 depends on VAR0. Each variable is \\ represented by a JSON string with keys in below.\end{tabular} \\ \hline
pair["name"]       & [str]        & Variable names.                                                                            \\ \hline
pair["line"]       & [int]        & Variable’s line number in trimmed code.                                                    \\ \hline
pair["col"]        & [int]      & Variable’s column number in trimmed code.                                                  \\ \hline
pair["operation"] &
  [str] &
  \begin{tabular}[c]{@{}l@{}}The operation performed on the variable. The value is either \\`w’ (representing a write operation) or `r’ (representing a read operation).\end{tabular} \\ \hline
\end{tabular}
\end{table*}
Creating the DRB-ML dataset is a multi-step process.
Firstly, we extract labels from each code snippet in the DRB dataset and store them in JSON.
Table \ref{tab:drbml_keys} illustrates the keys and their corresponding values in DRB-ML JSON files, each playing a crucial role in training ML models for data race detection.  The `data race-yes/no' label provides a binary indicator of data race conditions, while the `var\_pairs' label includes a list of variable pairs related to potential data races, along with their names, locations, and operation types (`w' for write or `r' for read). Together, these labels offer a comprehensive view of the features our models need to analyze for effective data race detection. This step is carried out using scripts that are designed to sift through code comments and metadata to find relevant information. Listing \ref{lst:drb193} shows an example in DRB-ML labels derived from a microbenchmark presented in Listing \ref{DRB001} from DRB. We omit the code content to better represent the paper. It is worth mentioning that the "line" value in DRB-ML is based on the code without comments.







\begin{lstlisting}[language=C, captionpos=b, caption=DRB001-antidep1-orig-yes.c, label={DRB001} ,firstnumber=47]
/*
A loop with loop-carried anti-dependence.
Data race pair: a[i+1]@64:10:R vs. a[i]@64:5:W
*/
#include <stdio.h>
int main(int argc, char* argv[])
{   
  int i;
  int len = 1000;

  int a[1000];

  for (i=0; i<len; i++)
    a[i]= i; 

#pragma omp parallel for
  for (i=0;i< len -1 ;i++)
    a[i]=a[i+1]+1;

  printf ("a[500]=%d\n", a[500] );
  return 0;
} 
\end{lstlisting}

\begin{lstlisting}[language=python, caption=DRB-ML-001.json, label=lst:drb193]
{
  "ID": "001",
  "name": "DRB001-antidep1-orig-yes.c",
  "DRB_code": "..."
  "data_race": 1,
  "trimmed_code": "..."
  "code_len": 262,
  "data_race_label": "Y1",
  "var_pairs": 
  [
      "{
         "name": ["a[i+1]", "a[i]"],
         "line": [64, 64],
         "col": [10, 5],
         "operation": ["R", "W"]
       }"
  ]
}
\end{lstlisting}
The second step involves the creation of a data template for the prompt-response pairs. The prompts are formulated to guide the LLM in identifying data races and to provide information about variables that might be causing them. The responses are simple labels indicating whether a data race exists or not.

In the final step, we employ scripts to pull the code and the information generated in the first step. The result is a structured prompt-response pair for each code in the DRB-ML dataset.

Upon completion of this process, each code snippet in the DRB-ML dataset contains three key pieces of information: the presence (or absence) of a data race, pairs of variables that could cause a data race, and the corresponding line numbers where these variables are found.

\begin{lstlisting}[language=python, caption=Prompt-response example for DRB-ML-193, label=prompt]
{
"prompt": """You are an HPC expert. Examine the following code and identify if there's a data race. If a data race is present, specify the variable pairs causing it, along with their line numbers and operations. Code: ...""",
"response": """Yes, the provided code exhibits data race issues. The data race is caused by the variable 'x' at line 9 and the variable 'x' at line 26. Both instances involve write operations."""
}
\end{lstlisting}

\subsection{Experiment Setup}
\textbf{Dataset:} As outlined in Figure \ref{fig:pipe}, we employ two strategies to evaluate the proficiency of LLMs in data race detection. First, we extract a subset of DRB-ML, ensuring that the data items have token sizes of less than 4k to accommodate the input sequence size limits of the selected LLMs. This sub-set consists of 198 out of the total 201 entries in DRB-ML. For the prompt engineering approach, we utilize the labels in the sub-set to assess the performance of the LLMs. Conversely, for fine-tuning the LLMs, we rely on the prompt-response pairs in DRB-ML for the fine-tuning process. The performance of the fine-tuned LLMs is then evaluated using the labels from the dataset.

\noindent
\textbf{Models}: 
We start our experiments by employing four pre-trained large language models for data race detection. The chosen models, including GPT-3.5-turbo, GPT-4, Llama2-7b, and StarChat-beta with 16 Billion parameters, represent a variety of architectures and are reputed for their performance on a range of tasks.

\subsection{Prompt Engineering for Data Race Detection}
\label{sec:prompt}
Prompt engineering is a key technique in harnessing the power of Large Language Models. It is a process wherein users tailor input prompts to elicit a particular response from the model. The goal is to craft prompts that effectively guide the model's responses in the desired direction. While it is hard to define the best prompt~\cite{zhou2022large}, a well-designed prompt can enable the model to provide insightful, precise, and contextually appropriate answers.

For the specific task of data race analysis using LLMs, we first delineated the expectations regarding their output in the following scenarios:
\begin{enumerate}[leftmargin=*]
    \item \textbf{S1. Data Race Detection:} Given a code snippet, LLMs are expected to decisively and concisely determine the presence of a data race.
    \item \textbf{S2. Identification of Data Race Variables:} LLMs should analyze the code to identify the variables responsible for the data race.
    \item \textbf{S3. Details on Data Race-related Variables:} LLMs ought to disclose pertinent information concerning each involved variable, including its name, its line number in the code, and the specific operation (either read or write) performed on it.
\end{enumerate}

With our goals outlined, we started with two basic prompts for data race detection. As an illustration, Listing \ref{lst:basic_p1} focuses on data race detection (S1), while Listing \ref{lst:basic_p2} instructs the LLMs to provide details on data race variables in conjunction with their data race detection findings (S1-3). Intriguingly, our preliminary experiments revealed a notable variance in the data race detection outcomes, shown in table \ref{tab:basic-compare}, when comparing the responses generated from GPT-3.5-turbo with the two basic prompts. 

\begin{lstlisting}[language=Python, caption=Basic Prompt 1 (BP1) template, label=lst:basic_p1]
"""
You are an expert in High-Performance Computing. Examine the code presented to you and ascertain if it contains any data races.
Begin with a concise response: either 'yes' for the presence of a data race or 'no' if absent.

{Code_to_analyze}
"""
\end{lstlisting}

\begin{lstlisting}[language=Python, caption=Basic Prompt 2 (BP2) template, label=lst:basic_p2]
"""
You are an expert in High-Performance Computing. Examine the code presented to you and ascertain if it contains any data races.
Begin with a concise response: either 'yes' for the presence of a data race or 'no' if absent.
detail each occurrence of a data race by specifying the variable pairs involved, using the JSON format outlined below:
{
"name": Names of each pair of variables involved in a data race.
"line": line numbers of the paired variables within the code.
"col": column number of the paird variables with in their line.
"operation_types": Corresponding operations, 'W' for write operation and 'R' for read operation.
}

{Code_to_analyze}
"""
\end{lstlisting}

\begin{table}[h]
\caption{Data race detection results of GPT-3.5-turbo using basic prompts 1 (BP1) and basic prompts 2 (BP2) shown in listing \ref{lst:basic_p1} and \ref{lst:basic_p2}, respectfully.}
\label{tab:basic-compare}
\begin{tabular}{|c|c|c|c|c|c|c|c|}
\hline
Prompts & TP & FP & TN & FN & Recall & Precision & F1    \\ \hline
BP1     & 66 & 55 & 43 & 34 & 0.660  & 0.545     & 0.597 \\ \hline
BP2     & 35 & 26 & 72 & 65 & 0.35   & 0.574     & 0.435 \\ \hline
\end{tabular}
\end{table}

The findings showcased in Table \ref{tab:basic-compare} suggest that multi-task prompts necessitate meticulous crafting in contrast to their simpler, more concise counterparts. This observation aligns with prior research in prompt engineering, where "greedy" prompts yielded sub-optimal performance ~\cite{zhou2022large}. Given these insights, we opted to refine our prompt engineering for data race detection based on Listing \ref{lst:basic_p1} while addressing the tasks of S2 and S3 through the fine-tuning approach discussed in Section \ref{sec:finetune}.


To enhance the quality of prompts for data race detection, we integrated insights from traditional tools and principles of concurrent programming. We crafted a prompt shown in Listing \ref{lst:p1} to explicitly instruct the LLMs to look for instances where two or more threads are simultaneously accessing the same memory location without proper synchronization, and at least one access is a write operation. Our preliminary results in table \ref{tab:basic-compare} show that a simple and concise prompt may be more efficient. Therefore, we broke the 
instruction in Listing \ref{lst:p1} into two prompts and executed them sequentially in a chat mode of the LLMs. This Chain-of-thoughts (COT) strategy introduced by Zhang et al.~\cite{zhang2022automatic} facilitates step-by-step thinking before answering a question, making each step simple and concise.

\begin{lstlisting}[language=Python, caption=Advanced Prompt 1 (AP1) for data race detection. AP1 extends BP1 by giving some details of data race detection including its definition and key analysis.,mathescape=true, label=lst:p1]
"""
You are an expert in High-Performance Computing (HPC). Examine the provided code to identify any data races based on data dependence analysis.
For clarity, a data race occurs when two or more threads access the same memory location simultaneously in a conflicting manner, without sufficient synchronization, with at least one of these accesses involving a write operation. It's crucial to analyze data dependence before determining potential data races.
Begin with a concise response: either 'yes' for the presence of a data race or 'no' if absent.
\end{lstlisting}

\lstset{language=Python, mathescape=true}

\begin{center} 

\begin{lstlisting}[language=Python, title=Chain1 in AP2. Chain1 guides the LLMs to check the data dependence in the given code., label=lst:p2_1]
"""You are an expert in High-Performance Computing (HPC). Analyze data dependence in the given code.

{Code_to_analyze}
"""
\end{lstlisting}
\vspace{-0.2em} 

\begin{lstlisting}[language=Python, title={Chain2 in AP2. With the output of Chain1 as a part of its input, Chain2 focuses on the data race detection task.}, label=lst:p2_2]
"""A data race occurs when two or more threads access the same memory location simultaneously in a conflicting manner, without sufficient synchronization, with at least one of these accesses involving a write operation. Identify any data races based on the given data dependence information.
Begin with a concise response: either 'yes' for the presence of a data race or 'no' if absent.
"""
\end{lstlisting}
\vspace{-1.5em} 
\captionof{lstlisting}{Advanced Prompt 2 (AP2). AP2 utilizes the chain-of-thoughts strategy to break AP1 into a chain. Chain1 and Chain2 are connected using LangChain's SequentialChain.}
\label{lst:p2}
\end{center} 



In summary, we employed various prompt engineering strategies for data race detection, referencing Listings \ref{lst:basic_p1}, \ref{lst:basic_p2}, \ref{lst:p1}, and \ref{lst:p2}. 




\subsection{LLM Fine-tuning for Data Race Analysis}
\label{sec:finetune}
\noindent
\textbf{Settings.}
The DRB-ML dataset, as detailed in Section \ref{sec:drbml}, provides foundational prompt-response templates designed specifically for data race detection. Building on this, we crafted two distinct prompt-response sets from the DRB-ML templates: one for detecting data races and another for identifying the associated variables. Our fine-tuning process follows prior works utilizing human feedback to enhance large language models~\cite{ziegler2019fine}. 

We chose the Llama2-7b and StarChat-beta models as our candidate base models for fine-tuning. We employed PyTorch version 2.01 and DeepSpeed 0.9.5 to support fine-tuning. For the Llama2-7b model, we adopted a learning rate of 2e-4, set the maximum sequence length to 256, and used the Adam optimizer. Conversely, for the StarChat-beta model, all settings remained consistent except for a learning rate adjustment to 9.65e-6.  We set the batch size to be 4 per GPU for training. To optimize memory usage during fine-tuning, we integrated QLoRA \cite{dettmers2023qlora}, setting the LoRA attention dimension to 64 and applying a dropout rate of 0.1. Our training process utilized the cross-entropy loss.

\noindent
\textbf{Fine-tuning objective. }Three scenarios were introduced in Section \ref{sec:prompt} for data race analysis. In the fine-tuning approach, we set two objectives for LLM fine-tuning: First, LLM fine-tuning for data race detection.
And second, LLM fine-tuning for data race variable identification.

\noindent
\textbf{Fine-tuning dataset. }Using the DRB-ML dataset, we utilized labels in the DRB-ML dataset to create two sets of 198 prompt-response pairs for data race detection and variable identification.  
\begin{itemize}
    \item Listing \ref{lst:sft_p1} shows an instance of prompt-response pairs derived from Listing \ref{lst:basic_p1} for LLM fine-tuning for basic data race detection.  
    \item Listing \ref{lst:sft_p2} shows an instance of prompt-response pairs derived from LLM fine-tuning for advanced data race detection with variable identification.
\end{itemize}



\begin{lstlisting}[language=python, caption=Instance of basic fine-tuning (basic-FT) prompt-response pairs for data race detection., label=lst:sft_p1]
{
"prompt":
"""
You are an expert in High-Performance Computing. Examine the code presented to you and ascertain if it contains any data races.
Begin with a concise response: either "yes" for the presence of a data race or "no" if absent.

{Code_to_analyze}
""",
"response": """yes"""

}
\end{lstlisting}

\begin{lstlisting}[language=Python, caption=Instance of advanced fine-tuning (advanced-FT) prompt-response pairs for advanced data race detection, label=lst:sft_p2]
{
"prompt":
"""
You are an expert in High-Performance Computing. Examine the code presented to you and ascertain if it contains any data races.
Detail each occurrence of a data race by specifying the variable pairs involved using the JSON format outlined below:
{
"variable_names": Names of each pair of variables involved in a data race.
"variable_locations": line numbers of the paired variables within the code.
"operation_types": Corresponding operations, either 'write' or 'read'.
}
{}
""",
"response":
"""
"yes",
{
    "data_race": 1,
    "variable_names": ["a[i]", "a[i+1]"],
    "variable_locations": [14, 14],
    "operation_types": ["write", "read"]
}
"""
}
\end{lstlisting}


\subsection{Five-fold Crossing Validation}
\label{sec:5folds}
We implemented a stratified k-fold cross-validation approach with $k=5$ to accomplish an unbiased evaluation.  This method is designed to retain a consistent proportion of positive to negative samples in each fold, mirroring the overall dataset's structure.

The subset of DRB-ML used in our work showcases a distribution of roughly 50.5\% positive(data race-yes) cases and 49.5\% negative(data race-no) cases. In crafting the 5-fold cross-validation, each fold is meticulously constructed to emulate this distribution. This delineation averages out to each fold, accommodating about 20 positive cases and 19.6 negative cases.

Given the indivisibility of data points in a practical setting, the allocation was determined as follows: Three of the folds were populated with both 20 positive and 20 negative cases, making up 40 data points in each of these folds. The remaining two folds were assembled with 20 positive cases and 19 negative cases each, resulting in 39 data points for each of these folds. By adopting this stratified 5-fold cross-validation, we provide a representative sample in each partition, ensuring a comprehensive and robust evaluation of LLMs.



\subsection{Evaluation Metrics}
In our study, we assess the performance of Large Language Models (LLMs) by examining their outputs in the context of three scenarios, as detailed in Section \ref{sec:prompt}. These scenarios—S1, S2, and S3—serve as binary classification tasks, allowing us to compute the counts of True Positives (TP), False Positives (FP), True Negatives (TN), and False Negatives (FN) by comparing the LLM outputs with the ground truth.

To quantify the performance of the LLMs, we utilize established metrics such as recall (R), precision (P), and the F1 score (F1). Additionally, we compute the average value (AVG) and standard deviation (SD) for these metrics across 5-fold cross-validation experiments.

It's worth noting that the values for recall, precision, and F1 score—as well as their respective averages—range from 0 to 1, with higher values indicating better performance. Additionally, a lower standard deviation is indicative of more consistent performance across the different folds of validation, making a lower SD score preferable.

\section{Experimental Results} 
This section presents the outcomes of our experiments, encompassing both prompt engineering and model fine-tuning exercises.

\subsection{Prompt Engineering for Data Race Detection}
Leveraging the trimmed code snippets from the DRB-ML dataset, we produced a set of three prompts for every code following the three prompt engineering strategies discussed in Section \ref{sec:prompt}. As suggested by the results in Table \ref{tab:basic-compare}, we did not adopt BP2 because the "greedy" prompts yield sub-optimal performance.

\begin{itemize}[leftmargin=*]
    \item \textbf{BP1:} Based on the template from Listing \ref{lst:basic_p1}. This prompt is succinct, directing LLMs straightforwardly to detect data races.
    \item \textbf{AP1:} Derived from Listing \ref{lst:p1}, this prompt instructs LLMs to emulate traditional tool methodologies, emphasizing data dependence analysis prior to ascertaining potential data races.
    \item \textbf{AP2:} Adopting the template from Listing \ref{lst:p2}, this variant separates the dual steps from AP1, adhering to a Chain-of-Thought design approach.
\end{itemize}

Subsequent to the model's output generation, we transformed these outputs into prediction labels. These predictions were then evaluated against the definitive "data\_race" labels found within DRB-ML. Comprehensive results of this assessment can be found in Table \ref{tab:llmres1}, where values in \textbf{bold} signify the best performance across all tools, while values in \textcolor{red}{red} denote the top-performing LLM.



\begin{table}[h]
\caption{Comparison of a representative traditional tool, Intel Inspector, and four LLMs: GPT-3.5-turbo, GPT-4, StarChat-beta, and Llama2-7b. We use three prompts: BP1, AP1, and AP2 to check if given codes contain data race. Values in \textbf{bold} signify the best performance across all tools, while values in green denote the top-performing LLM.}
\label{tab:llmres1}
\resizebox{\columnwidth}{!}{%
\begin{tabular}{|c|c|c|c|c|c|c|c|c|}
\hline
Model                     & Prompt   & TP & FP & TN & FN & R                       & P & F1    \\ \hline
Inspector & N/A
   &
  88 &
  {\color[HTML]{212121} 44} &
  53 &
  11 &
  \textbf{0.889} &
  0.667 &
  \textbf{0.762} \\ \hline
                           & BP1 & 66 & 55 & 43 & 34 & 0.660                        & 0.545     & 0.597 \\ \cline{2-9} 
                           & AP1 & 63 & 56 & 42 & 37 & 0.630                        & 0.529     & 0.575 \\ \cline{2-9} 
\multirow{-3}{*}{GPT3}     & AP2 & 69 & 54 & 44 & 31 & 0.690                        & 0.561     & 0.619 \\ \hline
                           & BP1 & 77 & 28 & 70 & 23 & 0.770                        & 0.733     & 0.751 \\ \cline{2-9} 
                           & AP1 & 78 & 30 & 68 & 22 & \textcolor{codegreen}{0.780} & 0.722     & 0.750 \\ \cline{2-9} 
\multirow{-3}{*}{GPT4} &
  AP2 &
  78 &
  28 &
  68 &
  22 &
  \textcolor{codegreen}{0.780} &
  \textcolor{codegreen}{\textbf{0.736}} &
  \textcolor{codegreen}{0.757} \\ \hline
                           & BP1 & 63 & 68 & 30 & 37 & 0.630                        & 0.481     & 0.545 \\ \cline{2-9} 
                           & AP1 & 62 & 67 & 31 & 38 & 0.620                        & 0.481     & 0.541 \\ \cline{2-9} 
\multirow{-3}{*}{StarChat} & AP2 & 63 & 61 & 37 & 37 & 0.630                        & 0.508     & 0.563 \\ \hline
                           & BP1 & 65 & 57 & 41 & 35 & 0.650                        & 0.533     & 0.586 \\ \cline{2-9} 
                           & AP1 & 65 & 57 & 41 & 35 & 0.650                        & 0.533     & 0.586 \\ \cline{2-9} 
\multirow{-3}{*}{Llama}   & AP2 & 66 & 55 & 43 & 34 & 0.660                        & 0.545     & 0.597 \\ \hline
\end{tabular}

}
\end{table}

\subsection{Basic LLM Fine-tuning: Data Race Detection}
To the best of our understanding, neither the training dataset for LLama2-7b nor StarChat-beta incorporates the DataRaceBench data upon reviewing their source. As such, we adopted the 5-fold cross-validation methodology detailed in Section \ref{sec:5folds} to fine-tune these open-source LLMs. We employed the basic prompt-response (basic-FT) pairs throughout the fine-tuning and validation phases, as exemplified in Listing \ref{lst:sft_p1}. 

Table \ref{tab:res-ft1} presents the results from this 5-fold cross-validation for the fine-tuned StarChat-beta and LLama2-7b models. The up-arrow indicates the performance increase by fine-tuned models compared with the original pre-trained versions. 

Broadly speaking, the fine-tuned models demonstrated enhanced F1 score and consistency performance. The StarChat-beta model registered improvements across nearly all metrics for data race detection, with the sole exception being recall consistency. Conversely, while the Llama2-7b model saw a dip in its recall score, it exhibited advancements in other evaluation metrics.

\begin{table}[h]
\caption{Average (AVG) and Standard Deviation (SD) of Recall, Precision, and F1 Score from a 5-fold cross-validation for data race detection using StarChat-beta, Llama2-7b, and their fine-tuned (FT) models with basic-FT prompts. Green indicates improved performance with fine-tuned models, while red signifies decreased performance.}
\label{tab:res-ft1}
\resizebox{\columnwidth}{!}{%
\begin{tabular}{|c|c|c|c|c|c|c|}
\hline
    Model     & AVG of R        & SD of R         & AVG of P        & SD of P         & AVG of F1      & SD of F1       \\ \hline
StarChat       & 0.630          & \textbf{0.045} & 0.482          & 0.041          & 0.546          & 0.039          \\ \hline
StarChat-FT    & \textcolor{codegreen}{\textbf{0.670}} & {\color[HTML]{CB0000} 0.057}          & \textcolor{codegreen}{0.541}    & \textcolor{codegreen}{\textbf{0.037}} & \textcolor{codegreen}{\textbf{0.598}} & \textcolor{codegreen}{\textbf{0.038}} \\ \hline
Llama    & 0.650          & 0.137          & 0.532          & 0.094          & 0.584          & 0.109          \\ \hline
Llama-FT & {\color[HTML]{CB0000} 0.640}          & \textcolor{codegreen}{0.082}         & \textcolor{codegreen}{\textbf{0.543}} & \textcolor{codegreen}{0.054}        & \textcolor{codegreen}{0.586}           & \textcolor{codegreen}{0.061}        \\ \hline
\end{tabular}%
}
\end{table}






\subsection{Advanced LLM Fine-tuning: Data Race Detection and Data Race Variable Identification}
As highlighted in the approach section, identifying data race-related variable pairs and extracting their detailed information poses significant challenges. Initially, we assessed the LLMs' performance concerning data race variable identification. Subsequently, we specifically fine-tuned the StarChat-beta and Llama2-7 models for this task.

\begin{table}[h]
\caption{Comparison of results of advanced data race detection with variable identification, using four LLMs. Values in bold signify the best performance across all models.}
\label{tab:llmres2}
\begin{tabular}{|c|c|c|c|c|c|c|c|}
\hline
Model & TP & FP & TN & FN & Recall         & Precision      & F1             \\ \hline
GPT3  & 12 & 54 & 44 & 88 & 0.120          & 0.182          & 0.145          \\ \hline
GPT4  & 14 & 31 & 67 & 86 & \textbf{0.140} & \textbf{0.311} & \textbf{0.193} \\ \hline
  StarChat  & 7  & 66 & 32 & 93 & 0.070          & 0.096          & 0.081          \\ \hline
  Llama  & 5  & 65 & 33 & 95 & 0.050          & 0.071          & 0.059          \\ \hline
\end{tabular}
\end{table}

Table \ref{tab:llmres2} showcases the performance metrics of the selected models before fine-tuning while Table \ref{tab:res-ft2} showcases the results from the 5-fold cross-validation.  The fine-tuned StarChat-beta and LLama2-7b models are compared to their original pre-trained versions. We consistently employed the advanced-FT prompt-response pairs throughout the fine-tuning and validation stages, as depicted in Listing \ref{lst:sft_p2}. Although the performance of the StarChat-beta model improved after fine-tuning, this enhancement came with an added inconsistency. Conversely, the Llama2-7b model didn't exhibit any significant improvements, potentially due to the limited training dataset.


\begin{table}[h]
\caption{Average (AVG) and Standard Deviation (SD) Recall, Precision, and F1 score of the 5-fold crossing validation for the advanced data race variable identification with StarChat-beta, Llama2-7b, and the fine-tuned (FT) models. Green indicates improved performance with fine-tuned models, while red signifies decreased
performance.}
\label{tab:res-ft2}
\resizebox{\columnwidth}{!}{%
\begin{tabular}{|c|c|c|c|c|c|c|}
\hline
    Model     & AVG of R        & SD of R         & AVG of P        & SD of P         & AVG of F1      & SD of F1       \\ \hline
StarChat    & \textbf{0.070} & \textbf{0.045} & 0.096          & \textbf{0.063} & 0.081          & \textbf{0.052} \\ \hline
StarChat-FT & \textbf{0.070} & { \color[HTML]{CB0000} 0.057 }
 & \textcolor{codegreen}{0.103}
 & { \color[HTML]{CB0000} 0.087 }
 & \textbf{\textcolor{codegreen}{0.083}}
 & { \color[HTML]{CB0000} 0.069 }
 \\ \hline
Llama    & 0.050          & 0.050          & \textbf{0.085} & 0.087          & 0.063          & 0.064          \\ \hline
Llama-FT & 0.050          & 0.050          & \textcolor{codegreen}{0.092} & \textcolor{codegreen}{0.086} & \textcolor{codegreen}{0.064} & \textcolor{codegreen}{0.063} \\ \hline
\end{tabular}
}
\end{table}





\subsection{Observation}
Through meticulously crafted experiments focused on data race analysis using LLMs, we derived the following insights from our results:
\begin{itemize}[leftmargin=*]
    \item In general, GPT-4 stands out as the premier pre-trained model for data race analysis, excelling particularly in identifying data race-related variables. Nevertheless, the open-source models, namely StarChat-beta and Llama2-7b, demonstrate significant potential. With the right fine-tuning, they could indeed surpass the GPT series in data race detection capabilities.
    \item While traditional tools achieve superior performance in terms of the F1 score when compared to LLMs, testing with the DataRaceBench data indicates that GPT-4 exhibits noteworthy potential. This is impressive, given that GPT-4 is designed for general-purpose tasks and not specifically optimized for this domain.
    \item Our initial results, showcased in Table \ref{tab:basic-compare}, indicate a clear trend: simple and concise prompts yield better results by LLMs. Our extensive prompt engineering results reinforce this observation, as presented in Table \ref{tab:llmres1}. Specifically, with the exception of the Llama2-7b model, all other models displayed enhanced performance with 'BP1'—a succinct prompt, as compared to 'BP2'—a multi-task oriented prompt, when it came to data race detection.
    \item Our results from fine-tuning demonstrate the potential of open-source LLMs in handling data race analysis tasks.

\end{itemize}

\subsection{Challenges and Possible Solutions}
In our exploration of data race analysis with LLMs, spanning from dataset preparation to LLM inference, fine-tuning, and evaluation, we encountered several challenges:

\begin{itemize}[leftmargin=*]
\item \textbf{Dataset:} The dataset preparation was both time-consuming and labor-intensive, further complicated by the scarcity of available datasets. This scarcity subsequently affected the efficacy of LLM fine-tuning. Potential remedies include:
\begin{itemize}
\item Crawling data from open-source repositories.
\item Generating synthetic datasets tailored for training.
\item Automating the dataset processing stages using LLMs.
\end{itemize}

\item \textbf{Natural Language Output Processing:} As text generation models, LLMs produce outputs in natural language. Parsing and processing these outputs present considerable challenges. One approach to mitigating this challenge is by directing LLMs to adhere to specific output formats. Initially, our DRB-ML dataset's prompt-response pairs, as exemplified in Listing \ref{prompt}, contained natural language outputs. We later transitioned to structured JSON outputs, as depicted in Listing \ref{lst:basic_p2}. Nonetheless, not every LLM consistently maintains designated output formats, leading us to employ regular expressions for parsing.

\item \textbf{General Challenges for PLP with LLMs:} Although LLMs have achieved great success in a lot of areas, their success happens mostly in NLP tasks. The processes of training data collection, tokenization, and embedding representations for the LLMs are all finely tuned to cater to the requirements of NLP applications.
Advancements in LLMs have recently incorporated programming language source codes and language-specific content into their training datasets. However, the quality of this training data remains suboptimal. A notable issue is the inclusion of incomplete or incorrect code snippets that cannot be successfully compiled by standard compilers. This deficiency has drawn our attention, highlighting the pressing need to enhance the quality of training data for Programmable Language Models in the context of programming tasks. Addressing this challenge is imperative to fully empower LLMs for effective performance in PLP tasks.
\end{itemize}


\section{Conclusion} 
In this paper, we have explored the capabilities of large language models for the task of detecting data races in OpenMP programs. A dedicated dataset, DRB-ML, was created based on DataRaceBench to evaluate and fine-tune LLMs. The results show that LLMs have the potential to become an alternative solution for data race detection. However, they cannot outperform traditional data race detection tools without improved training datasets or novel code representations that capture more code semantics.

In the future, we are interested in expanding DRB-ML to include more data items using data scraping and augmentation techniques. We will also explore different modalities beyond text as input, such as abstract syntax trees, dependence graphs, and control-flow graphs.

\begin{acks}
Prepared by LLNL under Contract DE-AC52-07NA27344(LLNL-CONF-853160) and supported by the U.S. Department of Energy, Office of Science, Advanced Scientific Computing Research.
This research was also funded in part by and used resources at the Argonne Leadership Computing Facility, which is a DOE Office of Science User Facility supported under Contract DE-AC02-06CH11357
\end{acks}

\bibliographystyle{ACM-Reference-Format}
\bibliography{correct23}










\end{document}